\documentclass[10pt,twocolumn,letterpaper]{article}

\usepackage{cvm}
\usepackage{times}
\usepackage{epsfig}
\usepackage{graphicx}
\usepackage{amsmath}
\usepackage{amssymb}

\usepackage{booktabs}
\usepackage{multirow}
\usepackage{tabularx}
\usepackage[misc]{ifsym}

\usepackage{algorithm}
\usepackage{algorithmic}
\usepackage{float}
\usepackage{pifont}
\usepackage[dvipsnames]{xcolor}
\usepackage{bbding}
\usepackage{textcomp, gensymb}
\usepackage{colortbl}
\usepackage{makecell}
\usepackage{balance}
\usepackage{caption}

\definecolor{mycyan}{cmyk}{.1,0,0,0}
\definecolor{mygray}{gray}{.95}
\definecolor{mypink}{rgb}{.99,.91,.95}
\definecolor{tabf}{rgb}{1, 0.7, 0.7}   
\definecolor{tabs}{rgb}{1, 0.85, 0.7} 
\definecolor{tabt}{rgb}{1, 1, 0.7}     
\newcommand{\cccf}[1]{\cellcolor{tabf}{\textbf{#1}}}
\newcommand{\cccs}[1]{\cellcolor{tabs}{#1}}
\newcommand{\ccct}[1]{\cellcolor{tabt}{#1}}

\newcommand{\algCmt}[1]{\hfill $\triangleright$ \text{#1}}

\usepackage[pagebackref=true,breaklinks=true,letterpaper=true,colorlinks,bookmarks=false]{hyperref}
\usepackage{cleveref}
\newcommand{\keywords}[1]{{\bf \emph{Keywords: #1}}}

\cvmfinalcopy

\def\cvmPaperID{****} 

\ifcvmfinal\fi
\begin{document}

\newcommand{\name}{LiDAR-GS}
\title{LiDAR-GS: Real-time LiDAR Re-Simulation using Gaussian Splatting}

\author{Qifeng Chen$^1$, Sicong Du$^1$, Tao Tang$^2$, Rengan Xie, Peng Chen$^1$, Yuchi Huo$^3$,Sheng Yang$^1$\textsuperscript{\Letter} \\
$^1$Unmanned Vehicle Dept. of CaiNiao Inc., Alibaba Group. \\$^2$Sun Yat-sen University. $^3$Zhejiang University. \\ \Letter \ denotes the corresponding author.
}

\twocolumn[{%
\renewcommand\twocolumn[1][]{#1}%
\maketitle
\begin{center}
    \centering
    \captionsetup{type=figure}
    \includegraphics[width=\textwidth]{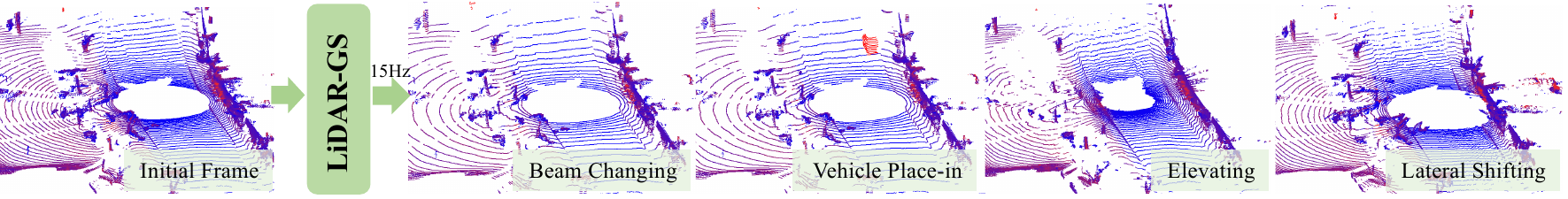}
    \captionof{figure}{We present {\name}, the LiDAR Gaussian Splatting for re-simulating LiDAR frames rendered in real-time.}
    \label{fig:teaser}
\end{center}%
}]

\begin{abstract}
We present LiDAR-GS, a Gaussian Splatting (GS) method for real-time, high-fidelity re-simulation of LiDAR scans in public urban road scenes. Recent GS methods proposed for cameras have achieved significant advancements in real-time rendering beyond Neural Radiance Fields (NeRF). However, applying GS representation to LiDAR, an active 3D sensor type, poses several challenges that must be addressed to preserve high accuracy and unique characteristics. Specifically, LiDAR-GS designs a differentiable laser beam splatting, using range-view representation for precise surface splatting by projecting lasers onto micro cross-sections, effectively eliminating artifacts associated with local affine approximations. Furthermore, LiDAR-GS leverages Neural Gaussian Representation, which further integrate view-dependent clues, to represent key LiDAR properties that are influenced by the incident direction and external factors. Combining these practices with some essential adaptations, e.g., dynamic instances decomposition, LiDAR-GS succeeds in simultaneously re-simulating depth, intensity, and ray-drop channels, achieving state-of-the-art results in both rendering frame rate and quality on publically available large scene datasets when compared with the methods using explicit mesh or implicit NeRF. Our source code is publicly available.
\end{abstract}


\section{Introduction}
\label{sec:intro}
There are four key distinctions between camera and LiDAR simulation:
(1) Most pinhole cameras share the common perspective projection model (even with varying distortions), whose local affine approximations in Normalized Device Coordinates (NDC) accumulates artifacts~\cite{huang2024ogs}, and thus not directly applicable for precise modeling of 360$^\circ$ rotating LiDAR sensors.
(2) Spherical Harmonics (SH)~\cite{Kerbl20233dgs}, designed for view-dependent color rendering, are suboptimal for intensity modeling of LiDAR. Compared to passive cameras, the variance of the incident direction and distance on the scanned surface would heavily influence the measurement.
(3) Also due to such an active sensing characteristic, some emitted laser beams would return no range measurement when the observed return signal has either too low amplitude or no clear peak, called ray-drop characteristics.
(4) Unlike rasterized camera pixels with uniform horizontal/vertical sampling, LiDAR is extremely sparse on the vertical FOV, typically only with 16-128 rotating laser beams.


Our approach addresses these four differences to bring-up LiDAR-GS accordingly:
(1) We choose range-view representation~\cite{kong2023rethinking} to organize laser beams and design a differentiable laser beam splatting method, projecting 3D Gaussians onto beam-aligned micro-planes instead of the unified NDC. This eliminates affine approximation errors while preserving geometric fidelity.
(2) For high-fidelity modeling and rendering on these LiDAR properties sensitive to the incident direction and distance, we propose to use Neural Gaussian Representation instead of the Spherical Harmonics, reaching higher quality of re-simulation according to our experiments (\Cref{tab:ablation} and \Cref{fig:ablation_mlp}).
(3) For simulating the ray-drop characteristics, we assign a learnable ray-drop probability to each Gaussian primitive, letting forward rendering to judge the validity of emitted laser beam.
(4) For vertical sparsity, we introduce spatial scale regularization and compact bounding-box strategies to constrain Gaussian scales and pixel coverage, ensuring precise gradient propagation. A dynamic decomposition mechanism further adapts the model to real-world driving scenarios without compromising real-time performance.


In conclusion, by combining methods and strategies mentioned above, we present the Gaussian Splatting approach for LiDAR re-simulation tasks called {\name}, which achieves state-of-the-art performance in both rendering quality and efficiency on publicly available datasets~\cite{sun2020waymo, liao2022kitti} compared to previous methods using explicit mesh or NeRF. Our solution also effectively addresses dynamic instances through decomposed reconstruction, seamlessly reintegrating them into the static background.
\section{Related Work}
\label{sec:rel_work}


Among applicable simulators~\cite{li2024choose}, CARLA~\cite{Dosovitskiy17} builds the traditional render engine upon manufactured virtual scenes with assets modeling, texture editing, and shading parameters tuning.
Such a high expense hinders a large-scale generation of virtual scenes.
Latest advancements such as MARS~\cite{wu2023mars} and UniSim~\cite{yang2023unisim} take advantage of NeRF's realistic rendering ability, enabling re-simulating sequences captured from real-world environments and guaranteeing a smaller sim-to-real gap~\cite{man2023towards,lindstorm2024sim2realgap}.
However, these NeRF-based simulators are comparably less efficient than GS-based rendering, requiring more effort to bring-up a real-time simulation system, and they neither support structuralized scene editing due to their implicit representation. Consequently, several approaches substitute the scene representation to GS for cameras (Section~\ref{sec:rel_work:cam}), whereas such a representation has not been thoroughly considered for LiDAR (Section~\ref{sec:rel_work:lidar}).


\subsection{Gaussian Splatting for Cameras}
\label{sec:rel_work:cam}

\textbf{Evolution of Scene Representation.} Using reconstruction methods to recover the driving scene for camera re-simulation has experienced three stages: Explicit reconstruction with ray-casting~\cite{dai2017bundlefusion}, NeRF~\cite{mildenhall2020nerf}, and GS~\cite{Kerbl20233dgs}. While rendering and Novel View Synthesis (NVS) quality is gradually advancing throughout these stages, both the efficiency and the asset editing ability drop in the NeRF stage. Thus, several sensor simulators~\cite{zhou2024hugsim} refactor their rendering pipeline by GS representation to support efficient rendering and flexible scene editing.

\textbf{Gaussian Splatting.} 3D Gaussian Splatting (3DGS)~\cite{Kerbl20233dgs}
proposes to rasterize a set of Gaussian ellipsoids to approximate the appearance of a 3D scene, achieving fast convergence of reconstruction. It also chooses to reversely splat intersected primitives onto rasterized sensor frames for acceleration, ensuring real-time rendering performance. 2DGS~\cite{Huang2DGS2024} replaces 3D Gaussians with 2D Gaussians for a more accurate ray-splat intersection, and employs a low-pass filter to avoid degenerated line projection. Scaffold-GS~\cite{lu2024scaffold} proposes to initialize a voxel grid and attach learnable features onto each voxel point and all attributes of Gaussians
are determined by interpolated features and lightweight neural networks, and Octree-GS~\cite{ren2024octree} further introduces a level-of-detail strategy to better capture details. GS++~\cite{huang2024ogs} analyzes the mathematical expectation of the 3DGS error function, and optimizes the projection to adapt to different camera models, such as fisheye. AbsGS~\cite{ye2024absgs} proposes the absolute value accumulation of gradients to further optimize the  growth and splitting strategies.

\subsection{LiDAR Re-simulation Methods}
\label{sec:rel_work:lidar}


\textbf{Explicit reconstruction with ray-casting.} 
Explicit reconstruction methods mainly use two types of 3D data structures: The first category, 3D mesh, is reconstructed by either a stand-alone point-to-mesh process such as SPSR~\cite{kazhdan2013spsr}, or Truncated Signed Distance Function (TSDF) Volumes~\cite{dai2017bundlefusion} with Marching Cubes~\cite{lorensen1987marchingcubes}. The second category, Surface Elements (Surfels), regards points as planar discs with additional normal and radius attributes, and gradually refine their attributes through scan integration~\cite{whelan2015elasticfusion,cao2020recon}. LiDARsim~\cite{manivasagam2020lidarsim} reconstructs scenes as a Surfel-based 3D mesh and applies a neural network to learn the ray-drop characteristics as a binary classification problem. PCGen~\cite{li2023pcgen} further presents First Peak Averaging (FPA) ray-casting and a surrogate model of ray-drop, achieving improvements in the domain gap between real and simulated LiDAR frames.

\textbf{NeRF based methods.}
Recent LiDAR re-simulation methods~\cite{tao2023lidarnerf, Wu2023dynfl, Huang2023nfl, tao2024alignmif, zheng2024lidar4d} extend the NeRF formulation for novel LiDAR view synthesis. Among them, LiDAR-NeRF~\cite{tao2023lidarnerf} and NFL~\cite{Huang2023nfl} firstly propose the differentiable LiDAR re-simulation framework, which treats the oriented LiDAR laser beams as a set of rays and render 3D points and intensities in a similar manner as RGB. DyNFL~\cite{Wu2023dynfl} and LiDAR4D~\cite{zheng2024lidar4d} extend to dynamic scenes. AlignMiF~\cite{tao2024alignmif} further proposes to incorporate and align multimodal inputs for better LiDAR re-simulation.
%
Through our experiments, we found refactoring LiDAR re-simulation with Gaussian Splatting follows the trend of camera re-simulation discussed in \cref{sec:rel_work:cam}.

\textbf{Contemporaneous methods.} Contemporaneous publicly available work includes GS-LiDAR~\cite{jiang2025gslidar} and LiDAR-RT~\cite{zhou2024lidarrt}. On the same sequences, our method achieves a reconstruction efficiency that is $7.1\times$ faster.

\section{Preliminaries}
\label{sec:method:pred}

\begin{figure*}[!ht] 
\centering 
\includegraphics[width=\linewidth]{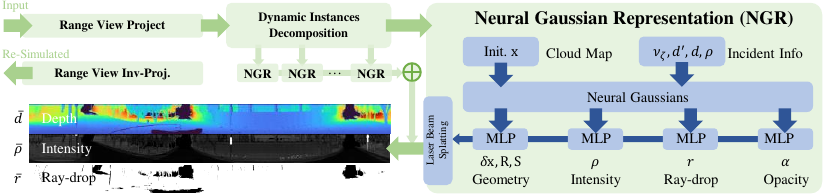}
\caption{\textbf{The overview of {\name} pipeline}. We choose to represent LiDAR frames as range-view images (\Cref{sec:method:pred:rangeview}), and use Gaussian Splatting to perform both reconstruction and rendering (\Cref{sec:method:pred:gs}). Specifically, we choose to use Neural Gaussian Representation (NGR) with four MLPs to learn final attributes for each Gaussian (\Cref{sec:method:lidar-gs}), where projecting and rasterizing these LiDAR Gaussians are performed through our proposed Laser Beam Splatting strategy (\Cref{sec:method:laser-beam-splatting}). Dynamic instances are separately reconstructed in their stand-alone coordinates for final stitching (\Cref{sec:method:dyn}).
}
\label{fig:overview}
\end{figure*}

\subsection{Range-view For LiDAR Representation}
\label{sec:method:pred:rangeview}

The range-view representation of LiDAR~\cite{kong2023rethinking,tao2023lidarnerf} counts image rows for the vertical FOV $f_v$, and columns for the horizontal FOV. During range-view conversion, each point \((x, y, z)\) in the sensor euclidean coordinate can be projected to a corresponding pixel location $(h, w)$ on the $\mathrm{H} \times 
\mathrm{W}$ range-view image, as:
\begin{equation}
\begin{pmatrix}h\\w\end{pmatrix}=\begin{pmatrix}\left(1-(\phi+f_\downarrow) / f_v \right)\\0.5 \cdot \left(1-\theta / \pi\right)\end{pmatrix} \cdot \begin{pmatrix} \mathrm{H} \\ \mathrm{W} \end{pmatrix},
\label{eq_lidar_rangeview_forward}
\end{equation}
where the vertical angle $\phi = \arcsin(z,\sqrt{x^2+y^2})$ and the horizontal angle $\theta = \arctan(y,x)$. The vertical FOV $f_{v} \triangleq f_\downarrow + f_\uparrow$ is divided into the lower and upper part from the $z=0$ horizontal plane.
%
The reverse operation of \Cref{eq_lidar_rangeview_forward} is:
\begin{equation}
\begin{pmatrix}
\phi\\\theta\end
{pmatrix}=
\begin{pmatrix}
f_\uparrow- f_v \cdot h / \mathrm{H} \\ \pi - 2 \pi \cdot w / \mathrm{W}
\end{pmatrix},
\label{eq_lidar_rangeview_backward}
\end{equation}
and the back-projection of \Cref{eq_lidar_rangeview_forward} can be operated as:
\begin{equation}
(x,y,z)=d \cdot \mathbf{d} \triangleq d \cdot (\cos\theta\cos\phi,\sin\theta\sin\phi, \cos\phi),
\label{rangeview_inverse_transformation}
\end{equation}
where $d$ stands for the distance along the ray to the scanned target, and the ray-direction $\mathbf{d}$ is determined by $(\phi, \theta)$. 

Through such projection, we can use the position \((h, w)\) in the range-view image to encode the corresponding information for \((x, y, z)\) in a multi-channel format, including depth, intensity, and ray-drop. Both $f_v$ and the exact vertical angle $\phi$ of each laser beam can be found in the LiDAR intrinsic parameters.

\subsection{3D Gaussian Splatting}
\label{sec:method:pred:gs}
3DGS~\cite{Kerbl20233dgs} uses Gaussian ellipsoids to explicitly represent the scene and introduces a point-based differentiable volume splatting. Specifically, the parameterization of the Gaussian primitives follows:
\begin{equation}
\mathbb{G}(\mathbf{x}) = \exp{(0.5 \cdot  \partial \mathbf{x}^\top{\Sigma}^{-1}\partial \mathbf{x})},
\label{eq_3dgs}
\end{equation}
where $\partial \mathbf{x}$ is the relative position from the center of the Gaussian ellipsoids, and ${\Sigma}$ is its 3D covariance matrix. The covariance matrix ${\Sigma}=\mathbf{R}\mathbf{S}\mathbf{S}^\top \mathbf{R}^\top$ is factorized into a scaling matrix $\mathbf{S}$ and a rotation matrix $\mathbf{R}$. To render an image, Gaussian ellipsoids will be projected into 2D image space, where the 2D projected covariance matrix ${\Sigma}'=\mathbf{J}\mathbf{W}{\Sigma} \mathbf{W}^\top \mathbf{J}^\top$ undergoes viewing transformation $\mathbf{W}$ and projective transformation, with $\mathbf{J}$ the Jacobian matrix of the local affine approximation during projection, designed to ensure that the ellipsoids maintain their regular shape after being projected from 3D to 2D. 
After the projection transformation, the color $\mathbb{C}(\mathbf{p})$ of each pixel $\mathbf{p}$ is computed by integrating overlapping Gaussian ellipses with $\alpha$-weighted:
\begin{equation}
\label{eq_3dgs_render}
\mathbb{C}(\mathbf{p})=\sum_{i \in \mathbf{N}_\mathbf{d}}c_i \cdot \alpha_{\xi_i}\prod_{j=1}^{i-1}(1-\alpha_{\xi_j}),
\end{equation}
where $c_i$ is the color from Spherical Harmonics~\cite{Kerbl20233dgs} of each Gaussian primitive $\xi_{i} \in \mathbf{N}_{\mathbf{d}}$ that is intersected by ray $\mathbf{d}$ emitted from the sensor viewpoint to the image pixel $\mathbf{p}$.



\section{LiDAR Gaussian Splatting}
\label{sec:method}

\subsection{\name{} Overview}
\label{sec:method:lidar-gs}

\textbf{Backward reconstruction.} The reconstructed scene is represented by a set of Gaussians $\bigcup \xi$, and each Gaussian primitive $\xi = \{\mathbf{x},\mathbf{R}, \mathbf{S}, \rho, r, \alpha\}$ stores position $\mathbf{x}$, rotation quaternion $\mathbf{R}$, scale coefficien $\mathbf{S}$, intensity $\rho$, ray-drop probability $r$, and opacity $\alpha$.

To enhance the fidelity of LiDAR depth and intensity, while incorporating the characteristics of this active sensor, we employ the Neural Gaussian Representation (NGR) motivated by NeRF~\cite{mildenhall2020nerf} and Scaffold-GS~\cite{lu2024scaffold}. Specifically, we indirectly utilize a scene feature vector $\mathbf{v}_\xi$ for each Gaussian (32-dim, initialized to zero), and use four lightweight Multi-Layer Perceptrons (MLPs) to obtain the desired attributes during rendering, as

\textit{$\mathbb{F}_{{\Sigma}}$, Offset and Covariance MLP: } 
We use $\mathbb{F}_{{\Sigma}}$ to learn the covariance components ${\Sigma}=\mathbf{R}\mathbf{S}\mathbf{S}^\top \mathbf{R}^\top$ and offset $\delta \mathbf{x}$ of init position $\mathbf{x}$, with inputs also related to view direction $\mathbf{d}'$:
\begin{equation}
    \mathbb{F}_{{\Sigma}} (\mathbf{v}_\xi, \mathbf{d}', d) \rightarrow \delta \mathbf{x}, \mathbf{R}, \mathbf{S},
\end{equation}
where we apply the offset $\delta \mathbf{x}$ as $\mathbf{x}=\mathbf{x}+\delta \mathbf{x} \cdot \mathbf{S}$~\cite{lu2024scaffold} during reconstruction. This facilitates the movement and expansion of the Gaussian in the direction of the gradient, promoting their growth and split progress. 

\textit{$\mathbb{F}_\rho$, Intensity MLP: }The definition of $\mathbb{F}_\rho$ is as follows:
\begin{equation}
    \mathbb{F}_\rho (\mathbf{v}_\xi, \mathbf{d}', d) \rightarrow \rho,
\end{equation} 
where we found MLP achieves better deduction than Spherical Harmonics when an input direction $\mathbf{d}'$ and the distance-of-flight $d$ is given (See \Cref{tab:ablation}).

\textit{$\mathbb{F}_{r}$, Ray-drop MLP: }
We additionally use $\mathbb{F}_{r}$ to learn about ray-drop probability, which means that laser beam do not return a value. The phenomenon of ray-drop usually exists in blind spots at close range and unmeasurable far range. Only in rare cases, the attenuation of laser caused by special materials can also exhibit ray-drop. Therefore, the input of the network contains information on view direction $\mathbf{d}'$, feature vector $\mathbf{v}_\xi$ for each Gaussian and distance $d$:
\begin{equation}
    \mathbb{F}_{r} (\mathbf{v}_\xi, \mathbf{d}', d) \rightarrow r.
\end{equation}

\textit{$\mathbb{F}_{\alpha}$, Opacity MLP: }
Similar with the previous method~\cite{Kerbl20233dgs}, we have also set up an opacity network $\mathbb{F}_{\alpha}$, as:
\begin{equation}
    \mathbb{F}_{\alpha} (\mathbf{v}_\xi, \mathbf{d}', d) \rightarrow \alpha, 
\end{equation}
where $\alpha$ interprets occlusions and invisibility among individual Gaussians, which is an indispensable attribute.

\textbf{Optimizing $\mathbf{v}_\xi$ and MLPs.} 
Given input real-world LiDAR scans, we use the following objective loss function to optimize the feature vector $\mathbf{v}_\xi$ of each Gaussian and four MLPs:
\begin{equation}
\begin{aligned}
\mathcal{L} &=\mathcal{L}_{d}+\mathcal{L}_{\rho}+\mathcal{L}_{r}+\mathcal{L}_{\mathbf{S}} + \mathcal{L}_{cd}, \\
\mathcal{L}_{\rho}&=(1-\lambda_\rho) \cdot \mathcal{L}_1+\lambda_\rho \cdot \mathcal{L}_\text{D-SSIM}, \\
\mathcal{L}_{\mathbf{S}}&=\frac{1}{N} \sum_{\xi_i}^{N} \mathrm{Prod}( \mathbf{S}_{\xi_i}), \\
\end{aligned}
\label{eq:loss}
\end{equation}
where the depth loss $\mathcal{L}_{d}$ follows $\mathcal{L}_1$ loss, and we set $\lambda_\rho = 0.2$ for the intensity loss $\mathcal{L}_{\rho}$. The CD loss $\mathcal{L}_{cd}$ follows the chamfer distance~\cite{cd} between the rendered point cloud and GT. The ray-drop probability loss $\mathcal{L}_{r}$ follows $\mathcal{L}_2$ loss, and we use a regularization loss $\mathcal{L}_{\mathbf{S}}$ to inhibit the enlargement of scale.

\textbf{Forward rendering.} 
Given a required free viewpoint and orientation for rendering, we choose to use the range-view rasterization and generate view-directions for interacting with splatted Gaussians.
Specifically, each pixel $\mathbf{p}$ in the rasterized range-view determines a view-direction $\mathbf{d}$ to fetch all related Gaussians $\mathbf{N}_\mathbf{d}$, and differential rendering for intensity $\rho$ presents an integration approach as $\bar{\rho} = \sum_{i \in \mathbf{N}_\mathbf{d}}\omega_{i}\rho_{i}$. 
By the same logic, the range-view image of depth and ray-drop probability can be calculated accordingly through the distance between the sensor and primitives. 
To determine the occurrence of ray-drop, we establish a threshold for the ray-drop probability, defined as $\bar{r} < \lambda_r$ ($\lambda_r = 0.5$). These criteria are employed to assess whether the laser beam is lost in a given ray direction. If so, clear all attributes for such a pixel.

\subsection{Laser Beam Splatting}
\label{sec:method:laser-beam-splatting}
In most pinhole camera GS methods~\cite{Kerbl20233dgs}, 3D Gaussians are splatted to the normalized image plane, employing the Jacobian matrix to approximate the non-linear projection process. Such a local affine approximation consequently degrades the quality of photo-realistic rendering~\cite{huang2024ogs}, and is also not applicable for LiDAR frames. Even in the range-view representation with equirectangular projection, the degradation still masters, because the LiDAR sampling is not equidistant in the vertical direction. Therefore, the differences between pinhole shutting for passive cameras and range-finding for active LiDAR sensors cannot be bridged by an single approximate projection.

\begin{figure}[ht] 
    \centering 
    \includegraphics[width=\linewidth]{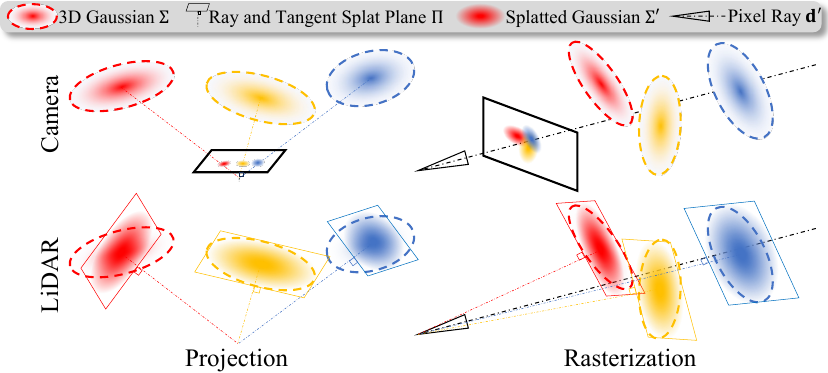}
    \caption{\textbf{Illustration of the proposed \emph{Micro Cross-Section Projection}}. Instead of splatting Gaussians to the normalized image plane like 3DGS~\cite{Kerbl20233dgs}, we calculate the cross-section micro-plane through the ray established between Gaussian mean and sensor viewpoint for splatting. During forward rendering, we collect acrossed Gaussians along the ray direction to integrate attributes for each pixel.} 
    \label{fig:proj} 
\end{figure}


In order to migrate Gaussian Splatting from passive camera to active LiDAR and minimize the distortion caused by the Gaussian projection and rasterization process, we introduce a new laser beam splatting method, called \emph{Micro Cross-Section Projection}, with implementation details illustrated in \Cref{fig:proj} and provided in \cref{alg:project_algorithm}.

\textbf{Gaussian projection.} We use the distance of flight $d = |\mathbf{x}-\mathbf{o}|$ (where $\mathbf{o}$ represents the sensor viewpoint) instead of z-buffering in the NDC space~\cite{Kerbl20233dgs} for depth buffering, and establish the target projection plane $\Pi$ perpendicular to the direction $\mathbf{d}'$ for each Gaussian, precisely located at its mean position $\mathbf{x}$. Hence, the projection of the Gaussian onto this plane $\Pi$ corresponds to the cross-section of the Gaussian itself, thus eliminating any distortion in the projection process.

Since there is no longer a normalized image plane or sphere to determine the range of affected pixels for each Gaussian during projection, we employed a compact Axis-Aligned Bounding-Box (AABB) to ascertain the pixel coverage of each Gaussian. Specifically, we establish the major axes $\mathbf{a}$ and minor axes $\mathbf{b}$ of the Gaussian projection on the projection plane and transform it into a rectangular bounding box $(b_h, b_w)$, where $b_h=\mathbf{b}/\tan\phi, b_w=\mathbf{a}/\tan\theta$. Due to the sparsity of the laser beam in the vertical FOV, the vertical angle $\phi$ is usually bigger than the horizontal angle $\theta$, leading to significant variations in gradient. Therefore, compared to directly using the maximum radius as the selection box~\cite{Kerbl20233dgs}, our strategy reduces the unnecessary pixel coverage, which is beneficial for optimizing the vertical direction. The ablation study in \Cref{tab:ablation} confirms the effectiveness of this strategy.

\begin{algorithm}[t]
\caption{Laser Beam Splatting}
\label{alg:project_algorithm}
\begin{algorithmic}[1]
\STATE \COMMENT{1. Gaussian Reconstruction}
\STATE \( \xi : \{\mathbf{x}, \mathbf{R}, \mathbf{S}, \rho, r, \alpha \} \)
\STATE \COMMENT{2. Gaussian Projection}
\FOR{ \( \xi \)}
    \STATE \( d \gets |\mathbf{x} - \mathbf{o}| , \ \ \mathbf{d}' \gets (\mathbf{x} - \mathbf{o}) / d \) \algCmt{Depth and Ray}
    \STATE \( n_1 \gets n_1 \perp \mathbf{d}', \ \ n_2 \gets n_2 \perp \mathbf{d}', n_1 \) \algCmt{Basis of plane}
    \STATE \( \Pi, \mathbf{J} \gets n_1, n_2 \) \algCmt{Tangent plane and Transformation}
    \STATE \( {\Sigma}' \gets \mathbf{J}\mathbf{W}{\Sigma} \mathbf{W}^\top \mathbf{J}^\top \) \algCmt{Projection of covariance}
    \STATE \( b_h, b_w \gets {\Sigma}' \) \algCmt{Range of affected pixels}
\ENDFOR
\STATE \COMMENT{3. Gaussian Rasterization}
\FOR{ \( \mathbf{d} \gets \mathbf{p} : (h, w) \)}
    \STATE \( \mathbf{N}_{\mathbf{d}} \gets \xi:\{b_h, b_w\} \) \algCmt{Depth buffering}
    \STATE \( \mathbf{p} : [\bar{\rho}, \bar{d}, \bar{r}] \gets [0, 0, 0] \)
    \algCmt{Render Target}
    \FOR{ \( \xi_i \in \mathbf{N}_{\mathbf{d}} \) }
        \STATE \( \epsilon \gets \arccos{(\mathbf{d}' \cdot \mathbf{d})} \)
        \STATE \( \partial\mathbf{x} \gets d,\epsilon \)
        \algCmt{Back-projection}
        \STATE \( \alpha_i \gets \alpha_{\xi_i} \cdot \mathbb{G}_i(\partial \mathbf{x}) \)
        \algCmt{Splatting and $\alpha$-weighting}
        \STATE \(  [\bar{\rho}, \bar{d}, \bar{r}] \gets \alpha_{i} \sum_{j=1}^{i-1}(1-\alpha_{j}) [\rho_{\xi_i}, d_{\xi_i}, r_{\xi_i}] \)
    \ENDFOR
    \IF{ \( \bar{r} < \lambda_r \) }
        \STATE \( \bar{d}, \bar{\rho} \gets 0 \) \algCmt{Ray-drop}
    \ENDIF
\ENDFOR
\end{algorithmic}
\end{algorithm}

\begin{figure}[ht] 
    \centering
    \includegraphics[width=0.8\linewidth]{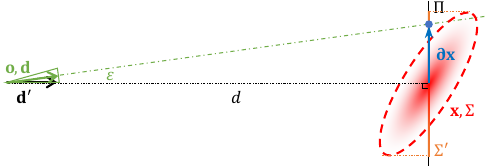}
    \caption{Sideview of the back-projection for each Gaussian during rendering.}
    \label{fig:inv_proj} 
\end{figure}

\textbf{Gaussian rasterization.} 
As shown in \Cref{fig:inv_proj}, during rendering of such a pixel with ray-direction $\mathbf{d}$, we back-project the relative position of the ray to the Gaussian onto the corresponding cross-section micro-plane $\Pi$. This step is crucial for calculating the distribution of the set of intersected Gaussians $\xi_i \in \mathbf{N}_{\mathbf{d}}$. Specifically for a pixel $\mathbf{p}$, if its corresponding ray $\{\mathbf{o},\mathbf{d}\}$ passes through a Gaussian at an actual distance of flight $d$, we pick the relative position $\partial \mathbf{x}$ as:
\begin{equation}
   \partial \mathbf{x} = d \cdot [ \mathbf{d} / (\mathbf{d}\cdot \mathbf{d}') - \mathbf{d}'].
\end{equation}

On the corresponding cross-section micro-plane $\Pi$, the real depth at the intersection of the ray and the intersected Gaussian can be calculated by $d/ (\mathbf{d}\cdot \mathbf{d}')$
Following \Cref{eq_3dgs}, the Gaussian splatting of $\xi_i$ w.r.t. pixel $\mathbf{p}$ is used to calculate the rendering weight $\alpha_i$ for \Cref{eq_3dgs_render}. Consequently, all attribute of each pixel are computed by integrating these intersected Gaussians with $\alpha$-weighted. 

\subsection{Dynamic Instances Decomposition}
\label{sec:method:dyn}
\begin{figure}[htbp]
    \centering
    \includegraphics[width=\linewidth]{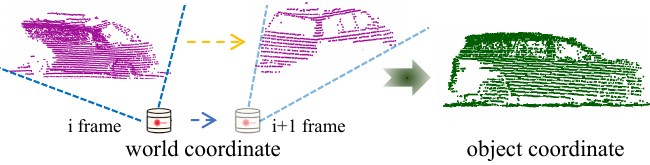}
    \caption{\textbf{Dynamic object decomposition and Gaussian modeling}. Given detection and segmentation results of each moving instance between consecutive frames, we separately stitch and reconstruct them in a standalone coordinate system for improving the completeness.}
    \label{fig:dynamic obj}
\end{figure}

In dynamic scenes, LiDAR-GS decomposes dynamic instances by utilizing detection results represented as bounding boxes, effectively distinguishing them from the static background. As illustrated in \Cref{fig:dynamic obj} and inspired by NSG~\cite{ost2021nsg}, each dynamic instance is reconstructed within a standalone coordinate system, and we use the Kabsch algorithm~\cite{lawrence2019kabsch} to estimate the relative pose between sequentially observed frames (If not provided). 
Using these tracked relative poses, we integrate observations from subsequent frames to enhance the completeness of instance-wise reconstruction. During rasterization, reconstructed Gaussians of each instance are placed w.r.t. the relative pose of the static background for a joint rendering.

\section{Experiments}
\label{sec:exp}

\subsection{Experimental Setting}
\textbf{Real-world Large Scene Datasets.} We verified the LiDAR-GS using 12 sequences from Waymo Open Dataset~\cite{sun2020waymo} and KITTI-360 Dataset~\cite{liao2022kitti}, which are public real-world datasets for autonomous driving scenarios. Consistent with previous LiDAR re-simulation methods~\cite{tao2023lidarnerf,Wu2023dynfl}, we utilized 50 continuous frames per sequence, designating every 10th frame for validation and the remaining frames for training. 
Please refer to \Cref{dataset} for the specific sequences chosen.

\begin{table}[H]
\caption{Sequences from Waymo and KITTI-360 for evaluation.}
\scriptsize
\centering
\setlength{\tabcolsep}{0.8mm}
\begin{tabular}{c|c}
\toprule
DATASET   & SCENE ID      \\ \midrule
\multirow{4}{*}{Static Waymo}  & 11379226583756500423\_6230\_810\_6250\_810   \\
& 10676267326664322837\_311\_180\_331\_180  \\
& 17761959194352517553\_5448\_420\_5468\_420   \\
& 1172406780360799916\_1660\_000\_1680\_000   \\ \midrule
\multirow{4}{*}{Dynamic Waymo} & 1083056852838271990\_4080\_000\_4100\_000    \\
& 13271285919570645382\_5320\_000\_5340\_000    \\
& 10072140764565668044\_4060\_000\_4080\_000    \\
& 10500357041547037089\_1474\_800\_1494\_800    \\ \midrule
\multirow{2}{*}{KITTI-360}      
& Seq 1538–1588, \ Seq 1728–1778   \\
& Seq 1908–1958, \ Seq 3353–3403    \\ \bottomrule
\end{tabular}
\label{dataset}
\end{table}

\textbf{Evaluation Metrics.} Following previous methods~\cite{tao2023lidarnerf,Huang2023nfl,tao2024alignmif,Wu2023dynfl}, our main evaluation metrics for depth including Chamfer Distance (CD), F-Score (using 5cm as threshold), Root Mean Squared Error (RMSE) and mean absolute error (MAE). Our evaluation metrics for intensity include Peak Signal-to-Noise Ratio (PSNR), Structural Similarity Index Measure (SSIM), Root Mean Squared Error (RMSE), and mean absolute error (MAE). 

\textbf{Baseline and Implementation.}
For static dataset, we used the official settings to conduct experiments with LiDAR-NeRF~\cite{tao2023lidarnerf} and AlignMiF~\cite{tao2024alignmif}. For NFL~\cite{Huang2023nfl} which have not yet released their source-code, we stripped the dynamic graph from DyNFL~\cite{Wu2023dynfl} and followed its settings to reproduce the results.
Additionally, we maintained the experimental settings as described in LiDAR-NeRF~\cite{tao2023lidarnerf} to reproduce the results on PCGen~\cite{li2023pcgen} and LiDARsim~\cite{manivasagam2020lidarsim}. 
For dynamics dataset, we reproduced the results using the same data sequences and official settings as DyNFL~\cite{Wu2023dynfl} and LiDAR4D~\cite{zheng2024lidar4d}, while also recording the official results of LiDARsim~\cite{manivasagam2020lidarsim} and UniSim~\cite{yang2023unisim} from DyNFL~\cite{Wu2023dynfl}. 
Meanwhile, we also compared our method with contemporaneous works, LiDAR-RT~\cite{zhou2024lidarrt} and GS-LiDAR~\cite{jiang2025gslidar}, by reproducing their results according to the official source code.

LiDAR-GS was trained for 5,000 iterations on RTX3090. 
We initialized 500,000 GS anchors, randomly sampled from all LiDAR frames and adopted the same growth and split strategy as Scaffold-GS~\cite{lu2024scaffold} and the same accumulative gradient strategy as AbsGS~\cite{ye2024absgs}. For densification, we adjusted the gradient threshold to 0.006, and set the final densification iterations to 3,000. The network structure employs double-layer 32-dims tiny MLP, implemented using tinycudnn~\cite{tiny-cuda-nn}.

\begin{table*}[!ht]
\caption{Average quantitative results for static scenes from the Waymo Open Dataset and KITTI-360.}
\small
\addtolength{\tabcolsep}{-6.1pt}
\begin{tabularx}{0.98\linewidth}{c|cccccccc|cccccccc}
\toprule
\multirow{3}{*}{Method} & \multicolumn{8}{c|}{Waymo}  & \multicolumn{8}{c}{KITTI-360}  \\ \cmidrule(r){2-17} 
& \multicolumn{4}{c|}{Point} & \multicolumn{4}{c|}{Intensity} & \multicolumn{4}{c|}{Point} & \multicolumn{4}{c}{Intensity} \\ \cmidrule(r){2-17} 
& \makecell{CD$\downarrow$}    & \makecell{F-score$\uparrow$} & \makecell{RMSE$\downarrow$}  & \multicolumn{1}{c|}{\makecell{MAE$\downarrow$}}  & \makecell{MAE$\downarrow$}    & \makecell{RMSE$\downarrow$}   & \makecell{PSNR$\uparrow$}  & \makecell{SSIM$\uparrow$} & \makecell{CD$\downarrow$}    & \makecell{F-score$\uparrow$} & \makecell{RMSE$\downarrow$}  & \multicolumn{1}{c|}{\makecell{MAE$\downarrow$}}  & \makecell{MAE$\downarrow$}    & \makecell{RMSE$\downarrow$}   & \makecell{PSNR$\uparrow$}  & \makecell{SSIM$\uparrow$}  \\ \midrule
LiDARsim~\cite{manivasagam2020lidarsim}          & 18.87 & 0.58    & 11.40 & \multicolumn{1}{c|}{4.75} & 0.074  & 0.12   & 18.56 & 0.32 & 1.04  & 0.66    & 6.04 & \multicolumn{1}{c|}{3.00} & \ccct{0.15}  & 0.20  & 13.64 & 0.09 \\
PCGen~\cite{li2023pcgen}             & 0.39  & 0.77    & 9.87  & \multicolumn{1}{c|}{3.07} & 0.072  & 0.12   & 18.29 & 0.33 & 0.26  & \ccct{0.84}    & 5.08 & \multicolumn{1}{c|}{2.43} & 0.16  & 0.22  & 13.11 & 0.06 \\
LiDARNeRF~\cite{tao2023lidarnerf}        & \ccct{0.19}  & \ccct{0.87}    & \ccct{7.52}  & \multicolumn{1}{c|}{\ccct{1.79}} & 0.039  & \cccs{0.07}  & \ccct{23.22} & \ccct{0.59} & \cccs{0.09} & \cccs{0.92}    & \ccct{3.70} & \multicolumn{1}{c|}{\ccct{1.35}} & \cccs{0.10}  & \ccct{0.15}  & \ccct{16.25} & \cccs{0.25}  \\
AlignMiF~\cite{tao2024alignmif}          & \cccs{0.16}  & \cccs{0.88}    & \cccs{7.39}  & \multicolumn{1}{c|}{\cccs{1.75}} & \ccct{0.038}  & \cccs{0.07}  & \cccs{23.27} & \cccs{0.60} & \cccs{0.09} & \cccs{0.92}    & \cccs{3.69} & \multicolumn{1}{c|}{\cccs{1.34}} & \cccs{0.10}  & \ccct{0.15}  & 16.19 & \cccs{0.25}  \\
NFL~\cite{Huang2023nfl}               & 0.27  & 0.84    & 8.40  & \multicolumn{1}{c|}{1.92} & \cccs{0.035}  & \ccct{0.08}  & 21.49 & 0.50 & \ccct{0.21}  & \ccct{0.84}    & 4.69 & \multicolumn{1}{c|}{1.59} & \cccs{0.10}  & \cccs{0.14}  & \cccs{16.61} & \ccct{0.18}  \\
\midrule
\rowcolor{mygray} LiDAR-GS          & \cccf{0.14}  & \cccf{0.91}    & \cccf{6.47}  & \multicolumn{1}{c|}{\cccf{1.46}} & \cccf{0.032}  & \cccf{0.06}  & \cccf{24.52} & \cccf{0.64} & \cccf{0.09}  & \cccf{0.92}    & \cccf{3.39} & \multicolumn{1}{c|}{\cccf{0.91}} & \cccf{0.08}  & \cccf{0.13}  & \cccf{17.38} & \cccf{0.49}  \\ \bottomrule
\end{tabularx}
\label{tab:avg_res}
\end{table*}

\begin{figure*}[!ht] 
\centering 
\includegraphics[width=\linewidth]{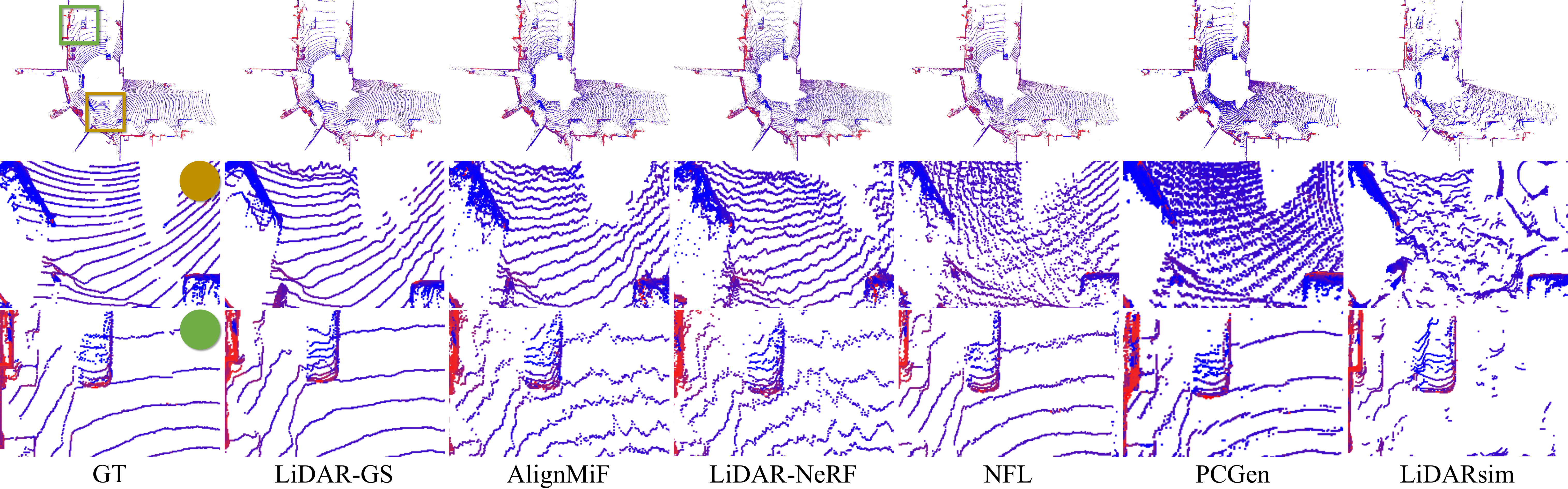}
\caption{For static scenes, qualitative comparison between {\name} and other methods on Waymo sequences. Points are colorized through their intensity from blue (0) to red (40).}
\label{fig:static_compare} 
\end{figure*}

\subsection{Main Results}

\begin{figure*}[!ht] 
\centering 
\includegraphics[width=\linewidth]{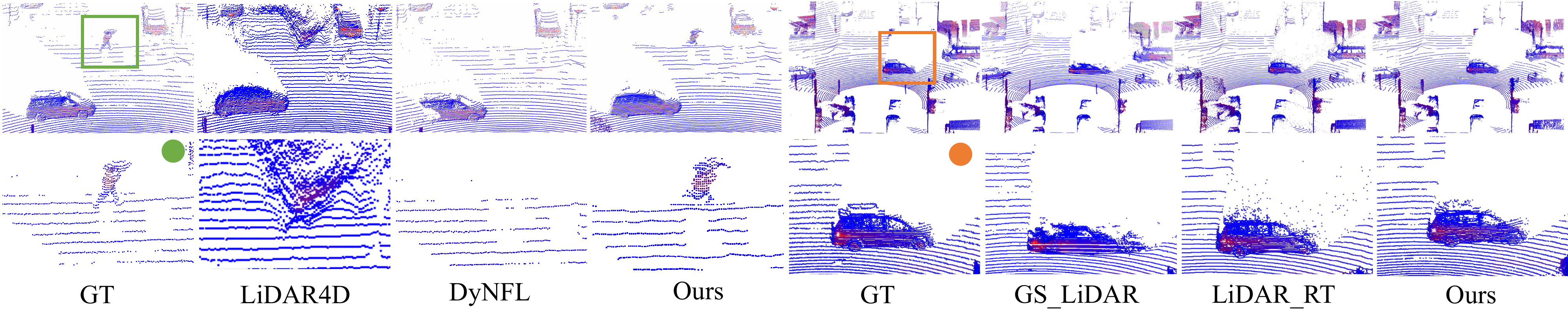}
\caption{For dynamic scenes, qualitative comparison between {\name} and other methods on Waymo sequence. Points are colorized through their intensity from blue (0) to red (40).}
\label{fig:dynfl_ours_comparision} 
\end{figure*}

\begin{table}[ht]
\caption{Average quantitative results for dynamic scenes.}
\centering
\scriptsize
\addtolength{\tabcolsep}{-4.6pt}
    \begin{tabularx}{0.98\linewidth}{c|cccc|cccc}
\toprule
 \multirow{2}{*}{Method}
&  \multicolumn{4}{c|}{Point}  & \multicolumn{4}{c}{Intensity} \\ \cmidrule(r){2-9} 
         & CD$\downarrow$ & F-scr$\uparrow$  & RMSE$\downarrow$  & MAE$\downarrow$  & MAE$\downarrow$    & RMSE$\downarrow$   & PSNR$\uparrow$  & SSIM$\uparrow$ \\ \midrule
LiDARsim~\cite{manivasagam2020lidarsim}   & 0.311 & 0.817 &  10.798 & 1.701 & - & - & - &- \\
UniSim~\cite{yang2023unisim}   & 0.143 & 0.902 &  2.631 & 0.256 & - & - & - &- \\
DyNFL~\cite{Wu2023dynfl} & 0.109 & 0.929 & 2.348 & 0.308 & 0.024 & 0.050 & 26.012 & 0.858 \\
LiDAR4D~\cite{zheng2024lidar4d} & 0.106 & 0.937 & 2.286 & 0.306 &\ccct{0.023} &\cccf{0.047} &\cccf{26.632} & \cccs{0.868} \\
LiDAR\_RT~\cite{zhou2024lidarrt} & \ccct{0.103} & \ccct{0.938} & \ccct{2.254} & \ccct{0.303} & \cccs{0.019} & \ccct{0.048} & \ccct{26.498} & \ccct{0.861} \\
GS\_LiDAR~\cite{jiang2025gslidar} & \cccs{0.096} & \cccs{0.940} & \cccs{2.150} & \cccs{0.300} & 0.024 & 0.050 & 25.431 & 0.840 \\
\midrule
\rowcolor{mygray}LiDAR-GS & \cccf{0.095} & \cccf{0.945} & \cccf{2.146} & \cccf{0.300} & \cccf{0.019} & \cccs{0.048} & \cccs{26.550} & \cccf{0.868} \\
\bottomrule
\end{tabularx}
\label{tab:avg_waymo_dynamic_scenes}
\end{table}

\begin{table}[ht]
\caption{Average training and inference costs. We reach real-time rendering for LiDAR re-simulation tasks.}
\centering
\setlength{\tabcolsep}{1.5mm}
\begin{tabular}{c|cc}
 \toprule
Method       & Train (min)$\downarrow$ & Inference (fps)$\uparrow$ \\ 
\midrule
LiDARsim~\cite{manivasagam2020lidarsim}   & \cccf{13}        & 0.33   \\
PCGen~\cite{li2023pcgen}      & \ccct{25}        & 0.01    \\
LiDAR-NeRF~\cite{tao2023lidarnerf} & 31  & \ccct{1.80}   \\
AlignMiF~\cite{tao2024alignmif}   & 110       & 0.10    \\
DyNFL~\cite{Wu2023dynfl}(NFL~\cite{Huang2023nfl}) & 464       & 0.15    \\
LiDAR4D~\cite{zheng2024lidar4d} & 426       & 1.7    \\
GS\_LiDAR~\cite{jiang2025gslidar} & 129       & 10.8    \\
LiDARRT~\cite{zheng2024lidar4d} & 213 & \cccf{20.7}    \\
\midrule
\rowcolor{mygray}  LiDAR-GS(w/o LBS)   & \cccs{17}      & \cccs{15.9}    \\ 
\rowcolor{mygray}  LiDAR-GS   & \cccs{18}      & \cccs{15.0}    \\ 
\bottomrule
\end{tabular}
\label{cost}
\end{table}

\Cref{tab:avg_res} and \Cref{tab:avg_waymo_dynamic_scenes} present our quantitative results on the public static dataset and dynamic dataset, respectively. As shown, our proposed method {\name} achieves best quantitative performance among most metrics. Meanwhile, \Cref{cost} demonstrates that our approach significantly reduces training and inference time compared to existing explicit reconstruction, NeRF methods and contemporaneous GS methods. This improvement marks a milestone that LiDAR re-simulation can now achiev the same or even higher frame rate than real-world sensors, typically at 10Hz, making large-scale re-simulation of real-world sequences feasible.

\textbf{Results on Static Dataset.} The quantitative comparison on the Waymo dataset with the static scene is displayed in \Cref{tab:avg_res}, illustrating LiDAR-GS has achieved state-of-the-art in most indicators. As shown in \Cref{fig:static_compare}, LiDAR-GS is superior to previous methods in modeling scene surfaces and intensity, such as roads, vehicles, and poles. We attribute this superior performance to our well-designed GS reconstruction process and efficient attribute prediction.


\textbf{Results on Dynamic Dataset.} The quantitative comparison on the Waymo dataset with dynamic objects is displayed in \Cref{tab:avg_waymo_dynamic_scenes}. We can find our proposed method achieves remarkable performance on most evaluation metrics, underscoring the realism of our re-simulated LiDAR scans. Although LiDAR-GS scores lower than LiDAR4D on some intensity metrics (attributed to its two-stage convolutional network optimization), our rendered results, as shown in \Cref{fig:dynfl_ours_comparision}, appear more realistic. And our reconstruction efficiency is significantly higher than that of GS-LiDAR, even though the performance of geometric accuracy is comparable.

\begin{figure*}[!ht] 
\centering 
\includegraphics[width=\linewidth]{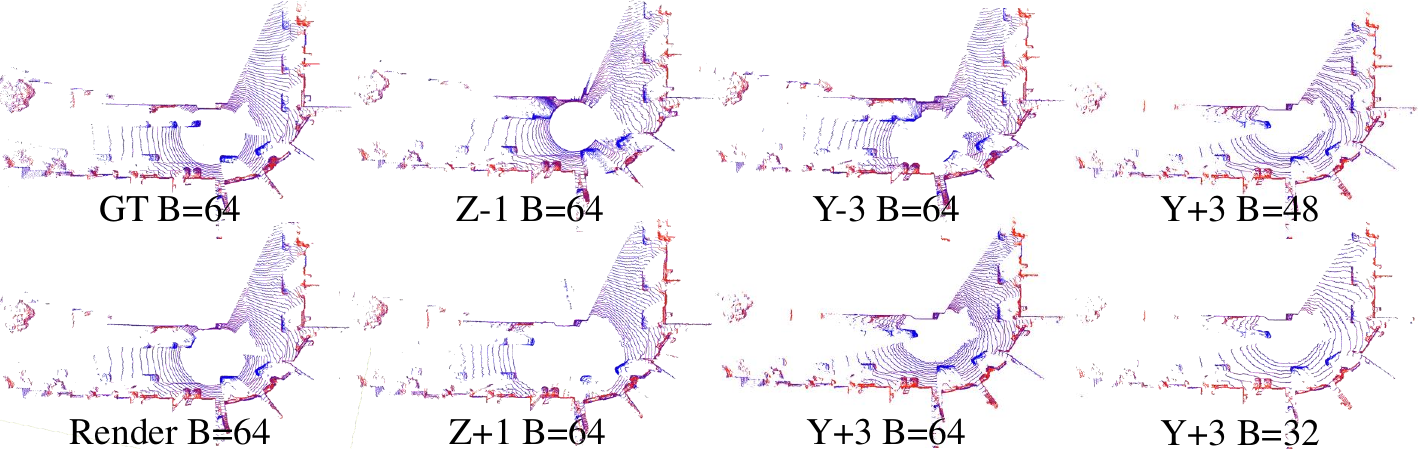}
\caption{Novel View Synthesis application when re-simulating the LiDAR sensor. `Y',`Z' stands for left, and up, respectively. Positional offset unit is performed in meters. `B' for number of laser beams to launch re-simulation. }
\label{fig:novel_lidar} 
\end{figure*}

\textbf{Results on Novel View Synthesis.} We adjusted the free-viewpoint by lateral shifting (Y) and elevating (Z), or change the number of laser beams (from 64 to 32) for LiDAR re-simulation on novel views as shown in \Cref{fig:novel_lidar}, demonstrating our effectiveness of NVS in driving scenarios. And supplementary video shows the more results of the re-simulation application, including \textbf{vehicle place-in, novel view, beam changing and simulated driving}.

\subsection{Ablation Results}

\begin{table}[ht]
\caption{Ablations for the proposed method on Waymo1067.}
\centering
\scriptsize
\addtolength{\tabcolsep}{-4.2pt}
    \begin{tabularx}{0.98\linewidth}{c|cccc|cccc}
\toprule

\multirow{2}{*}{Module} & \multicolumn{4}{c|}{Point} & \multicolumn{4}{c}{Intensity} \\\cmidrule(r){2-9} 
                & CD$\downarrow$     & F-score$\uparrow$ & RMSE$\downarrow$  & MAE$\downarrow$   & MAE$\downarrow$    & RMSE$\downarrow$   & PSNR$\uparrow$   & SSIM$\uparrow$   \\ \midrule
w/o NGR         & 3.907  & 0.243   & 12.1 & 6.64 & 0.041 & 0.074 & 22.57 & 0.60 \\ 
w/o LBS &   0.264 & 0.860 & 7.77 & 1.76 & 0.029 & \ccct{0.056} & 25.03 & 0.70 \\
w/o $\mathcal{L}_{\mathbf{S}}$ & 0.215  & 0.877   & \cccf{7.25} & \ccct{1.61} & \ccct{0.028}  & 0.057 & 25.40  & 0.71  \\
w/o AABB       & 0.219  & \ccct{0.879}  & 7.42  & 1.65  & 0.029  & 0.057 & \ccct{25.43}  & \ccct{0.71} \\
w/o   $\mathcal{L}_{cd}$    & \cccs{0.199} & \cccs{0.883}   & \ccct{7.28} & \cccs{1.56} & \cccf{0.027}  & \cccs{0.052} & \cccs{25.51}  & \cccs{0.73}  \\
\midrule
MLP w/o $\mathbf{v}_\xi$       & 0.221 & 0.871  & 7.52  & 1.66  & 0.029  & 0.058 & 25.35 & 0.69 \\
MLP w/o $d$     & \ccct{0.205} & 0.878  & 7.35  & 1.61  & \cccs{0.028}  & 0.058 & 25.38 & 0.70 \\
\midrule
\rowcolor{mygray} \name{}      & \cccf{0.194} & \cccf{0.885}   & \cccs{7.27} & \cccf{1.56} & \cccf{0.027}  & \cccf{0.052} & \cccf{25.54}  & \cccf{0.73}  \\ \bottomrule
\end{tabularx}
\label{tab:ablation}
\end{table}

\begin{figure}[ht] 
\includegraphics[width=\linewidth]{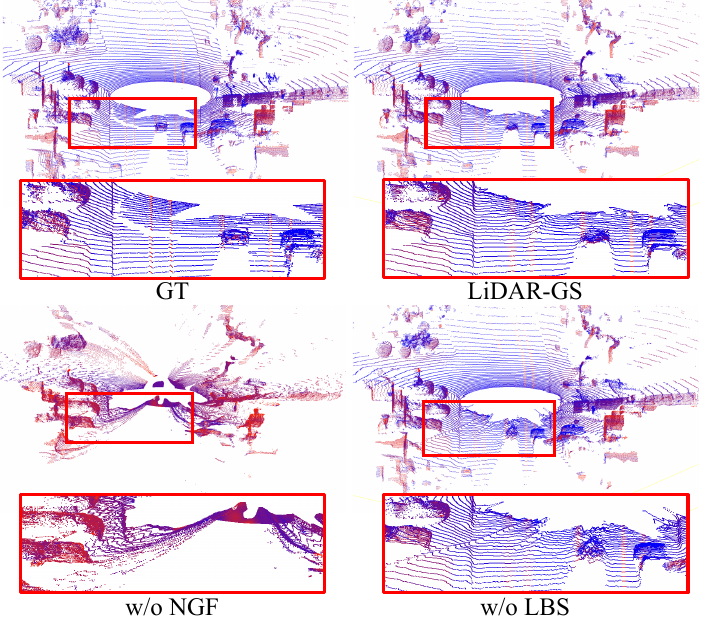}
\caption{Ablations on Neural Gaussian Representation and Laser Beam Splatting. }
\label{fig:ablation_mlp} 
\end{figure}

\textbf{Ablations on Neural Gaussian Representation.} We conducted an ablation experiment to remove the NGR. \Cref{tab:ablation} supports our assertions in \Cref{sec:method:lidar-gs}. The introduction of NGR significantly enhances the capability of scene representation. It can effectively consider the impact of incident direction and distance. As shown in \Cref{fig:ablation_mlp}, if Gaussian properties are not related to above factors, it cannot identify positions of ray-drop, leading to incorrect optimization of intensity and depth, such as road blind spots.

\begin{figure}[!ht] 
\centering 
\includegraphics[width=\linewidth]{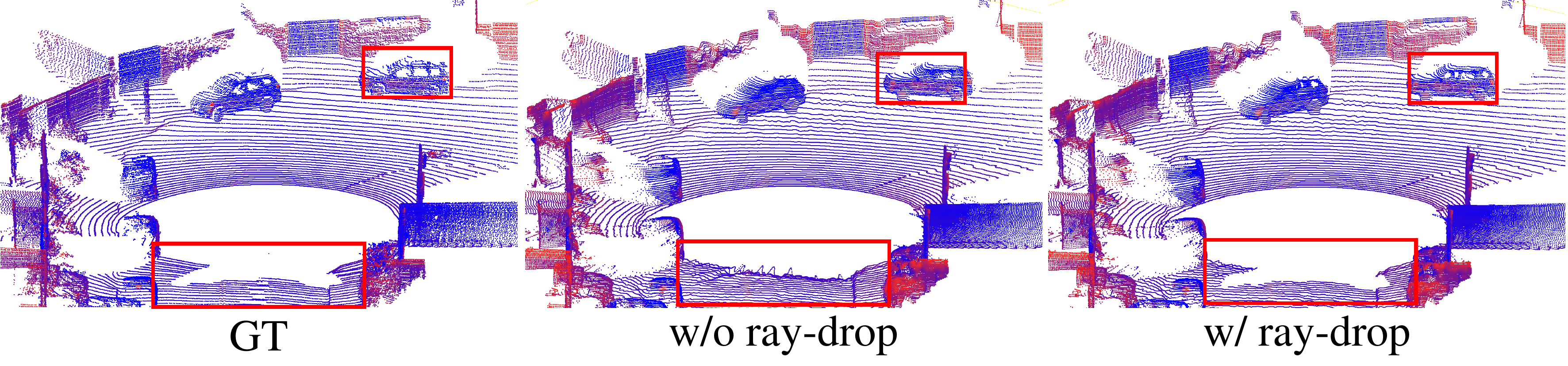}
\caption{Ablations on the necessity of adding ray-drop estimation. }
\label{fig:raydrop} 
\end{figure}

\textbf{Ablations on Laser Beam Splatting.}
We conducted an ablation experiment to remove the LBS. Specifically, we referred to the camera projection method of the vanilla 3DGS~\cite{Kerbl20233dgs} and projected Gaussian onto unified pano projection planes for rasterization. \Cref{tab:ablation} demonstrates that LBS is more suitable for LiDAR applications. The improvements provided by LBS are illustrated in \Cref{fig:ablation_mlp}, confirming our assertions in \Cref{{sec:method:laser-beam-splatting}}. Additionally, we implemented LBS using CUDA, resulting in only slight reduction in efficiency. The costs of projections are detailed in \Cref{cost}.

\textbf{Ablations on AABB.} The results in \Cref{tab:ablation} demonstrate that avioding unnecessary pixel coverage by using AABB for Gaussian projection is beneficial for further convergence.

\textbf{Ablations on regularization terms.} We also set up ablation experiments on the regularization term $\mathcal{L}_{\mathbf{S}}$ and loss $\mathcal{L}_{cd}$ mentioned in \Cref{eq:loss}. The depth rendering is an important basis for reflecting the ability of re-simulation, and introducing these two constraints brings some benefits to the Neural Gaussian Representation.

\textbf{Analysis of ray-drop properties.} \Cref{fig:raydrop} intuitively shows the effect of ray-drop properties on LiDAR re-simulation. Ray-drops are ubiquitous, some of which have certain regularities (such as blind spots), while others are affected by the incident direction and the environment, and it is difficult to find the distribution pattern intuitively. Therefore, parameterizing and optimizing ray-drops is helpful to improve LiDAR simulation capabilities.

\textbf{Analysis of MLP.} We conducted two additional experiments to demonstrate the necessity of incorporating additional inputs into the MLP network, including distance $d$ and feature vectors  $\mathbf{v}_\xi$. These inputs further assist the network in extracting features and producing accurate GS attributes.

\textbf{Discussion on the gap between real and simulation.} To gauge the gap between real and simulated scans, we employed detection model, INT~\cite{xu2022int}, on the same trajectory of dynamic data to evaluate the difference in detection results between two point cloud detection results relative to manually annotated GT. \Cref{tab:Object detection results} presents that this difference is negligible in this task, where mAP(mean Average Precision), mF1(mean F1 Score) and mAVE(mean Average Velocity Error) are used as detection metric.

\begin{table}[!ht]
\caption{Objects detection using simulation data.}
\centering
\scriptsize
\addtolength{\tabcolsep}{3.6pt}
    \begin{tabularx}{0.97\linewidth}{c|cccc}
     \toprule
    Data     & mAP$\uparrow$ & mF1$\uparrow$ & mAVE\_vx$\downarrow$ & mAVE\_vy$\downarrow$ \\ 
    \midrule
    Real     & 87.701 & 87.512 & 0.314 & 0.165 \\ 
    LiDAR-GS & 87.673 & 87.387 & 0.314 & 0.167 \\ 
    \bottomrule
\end{tabularx}
\label{tab:Object detection results}
\end{table}

\section{Conclusion}
In this paper, we present the LiDAR Gaussian Splatting method for real-time high-fidelity re-simulation of sensor scans in real-world urban road scenes. LiDAR-GS choose range-view representation to organize laser beams, and innovatively proposes a differentiable laser beam splatting based on micro cross-section plane. Furthermore, LiDAR-GS precisely simulates LiDAR properties using Neural Gaussian Representation and introducing some strategies dedicated to Gaussian laser beam, including AABB on projection constraints, and spatial scale regularization. LiDAR-GS has been verified in multiple dynamic and static scenes, and performs well. In summary, this GS-based rendering LiDAR re-simulation method will accelerate the progress of autonomous driving in data closure.

\textbf{Limitation and Future Work}. Due to the single training perspective and the point density that inversely proportional to the scanning distance, our proposed {\name} still exhibits degradation in occluded and extrapolated regions at far distances. Throughout our proposed {\name} pipeline, we also find several points worth further adjustments as a secondary priority:
Leverage the ability of generative models, such as diffusion model, to address depth occlusion caused by sparse perspective. And by fusing multi-modal sensor information, such as cameras, model can enhance the understanding of scene features.


{\small
\bibliographystyle{cvm}
\bibliography{main}
}

\end{document}